\documentclass[letterpaper, 10 pt, conference]{ieeeconf}  

\IEEEoverridecommandlockouts                              

\usepackage{graphicx} 
\usepackage{amsmath} 
\usepackage{amssymb}  
\usepackage{amsfonts}  
\usepackage{siunitx} 

\usepackage[linesnumbered]{algorithm2e}
\RestyleAlgo{ruled}
\newcommand{\eg}{\textit{e.g.,}~}
\newcommand{\ie}{\textit{i.e.,}~}
\DeclareMathOperator*{\argmin}{arg\,min}

\newcommand*{\sref}[1]{\S\ref{#1}}            
\newcommand*{\aref}[1]{Algorithm~\ref{#1}}  
\newcommand*{\tref}[1]{\text{TABLE~\ref{#1}}}   
\newcommand*{\fref}[1]{\text{Fig.~\ref{#1}}}
\newcommand*{\eref}[1]{\text{Eq.(\ref{#1})}}     

\usepackage[nobreak]{cite}

\usepackage[table]{xcolor}
\newcolumntype{C}[1]{>{\centering\arraybackslash}p{#1}}

\usepackage{caption}
\usepackage{subcaption}

\usepackage{hyperref}

\title{\LARGE \bf
Robust Instant Policy: Leveraging Student's t-Regression Model for Robust In-context Imitation Learning of Robot Manipulation
}

\author{Hanbit Oh$^{\dagger}$, Andrea M. Salcedo-V\'azquez, Ixchel G. Ramirez-Alpizar, and Yukiyasu Domae
\thanks{*This work was supported by BRIDGE SIGRoLe project}
\thanks{$^{\dagger}$ Corresponding author}
\thanks{
The authors are affiliated with the Industrial Cyber-Physical Systems Research Center at the National Institute of Advanced Industrial Science and Technology (AIST), Japan. }%
\thanks{© 2025 IEEE.  Personal use of this material is permitted.  Permission from IEEE must be obtained for all other uses, in any current or future media, including reprinting/republishing this material for advertising or promotional purposes, creating new collective works, for resale or redistribution to servers or lists, or reuse of any copyrighted component of this work in other works.}
}

\begin{document}

\maketitle
\thispagestyle{empty}
\pagestyle{empty}

\begin{abstract}
Imitation learning (IL) aims to enable robots to perform tasks autonomously by observing a few human demonstrations. Recently, a variant of IL, called In-Context IL, utilized off-the-shelf large language models (LLMs) as instant policies that understand the context from a few given demonstrations to perform a new task, rather than explicitly updating network models with large-scale demonstrations. However, its reliability in the robotics domain is undermined by \textit{hallucination} issues such as LLM-based instant policy, which occasionally generates poor trajectories that deviate from the given demonstrations. To alleviate this problem, we propose a new robust in-context imitation learning algorithm called the robust instant policy (RIP), which utilizes a Student’s t-regression model to be robust against the hallucinated trajectories of instant policies to allow reliable trajectory generation. Specifically, RIP generates several candidate robot trajectories to complete a given task from an LLM and aggregates them using the Student's t-distribution, which is beneficial for ignoring outliers (\ie hallucinations); thereby, a robust trajectory against hallucinations is generated. Our experiments, conducted in both simulated and real-world environments, show that RIP significantly outperforms state-of-the-art IL methods, with at least $26\%$ improvement in task success rates, particularly in low-data scenarios for everyday tasks. Video results available at \url{https://sites.google.com/view/robustinstantpolicy}

\end{abstract}
\begin{figure}[t!]
    \centering
    \includegraphics[width=0.5\textwidth]{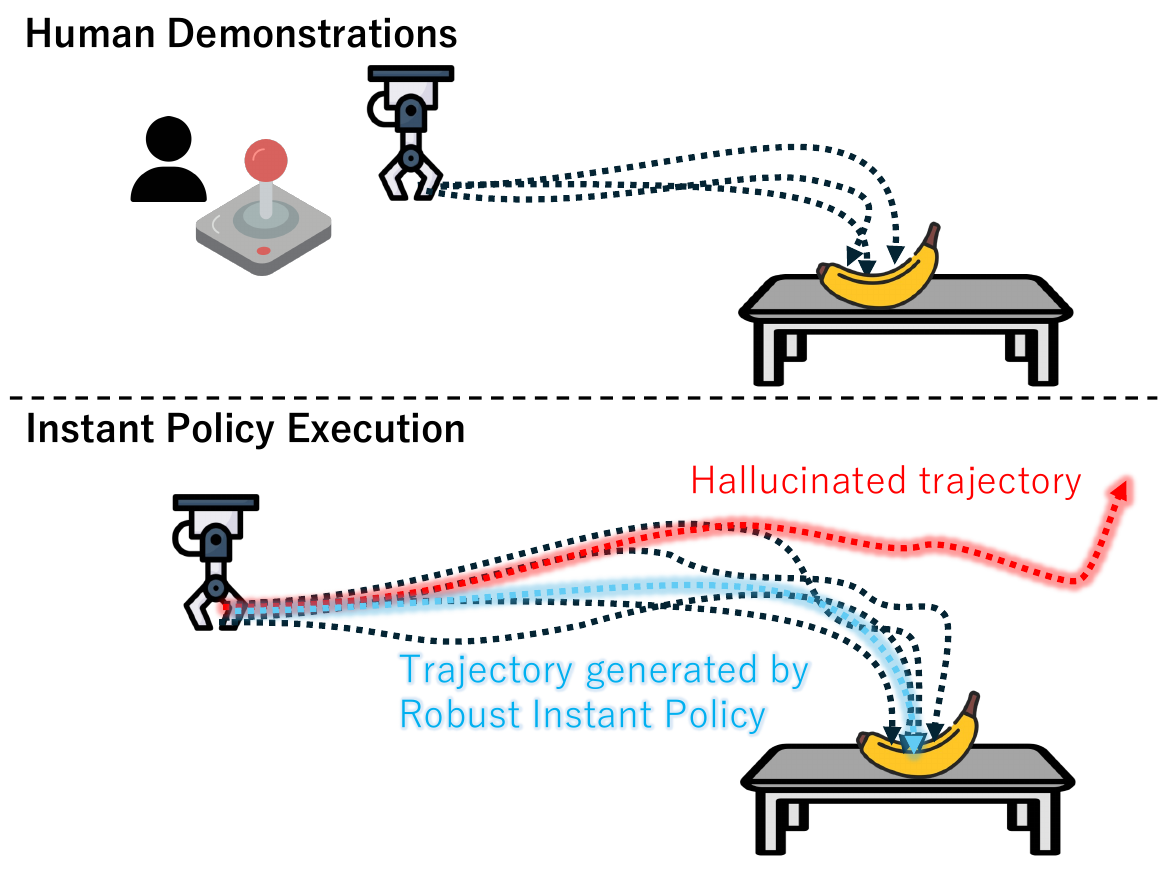}
    \caption{
        Robust instant policy (RIP) for a robotic banana-picking task.
        Given a few human demonstrations, LLM-based instant policy can capture the task's context and generate trajectories such as demonstrations, but some may deviate from demonstrations (red) owing to LLM's hallucinations. 
        In contrast, RIP generates a robust trajectory (blue) to hallucinations using a Student's t-regression model that averages trajectories of the instant policy while ignoring hallucinations.
    }
    \label{fig:idea}
\end{figure}

\section{INTRODUCTION}\label{sec:int}
Imitation learning (IL) is a promising technique for learning policies to automate robot manipulation by observing human demonstrations \cite{osa2018il}. It has shown considerable success in a wide range of applications with large datasets and highly expressive policy models \cite{o2024open, black2024pi_0, kim2024openvla}. However, despite its capabilities, it remains limited because it requires thousands of demonstrations and/or long-term model weight tuning to be applied to new tasks or environments. This motivates a more efficient paradigm for learning an instant policy that can be immediately adapted and deployed to new tasks while minimizing costs.

A previous study attempted to achieve this by in-context imitation learning (ICIL), in which a large transformer model trained on a variety of datasets can be immediately generalized to a new task by providing a few demonstrations of the new task as a context without updating the model \cite{fu2024contextil}. Although this technique is based on findings in the language and visual domains \cite{brown2020incontext}, where sufficient large-scale data are available, the amount of data in the robotics domain remains insufficient for widespread applications. In contrast, keypoint action tokens (KAT) translate robot trajectory data into keypoint-based text, enabling the off-the-shelf large language model (LLM) to be reused as an instant policy \cite{dipalo2024kat}, which has shown comparable task achievement to state-of-the-art IL methods by using fewer than 10 human demonstrations of a new task without updating the model.

Although the LLM is a useful tool for instant robotic agents, it faces the inherent issue of \textit{ hallucinations }, where its responses may be out of a given context and lack accuracy \cite{ji2023hallucination}. Specifically, this issue is critical in the robotics domain, where even a small loss of precision in robot trajectory generation owing to hallucinations can lead to task failure. Although several methods have been proposed to mitigate these hallucinations, they mainly focus on discrete language data \cite{ren2023robots} and are unsuitable for continuous robot trajectory data. Thus, an approach that uses continuous robot data is required to enhance the robustness of LLM-based instant robot policies.

Therefore, this study proposes a novel robust ICIL method that generates a robust trajectory against hallucinations of LLM through stochastic treatment (\fref{fig:idea}): the robust instant policy (RIP). In RIP, a few demonstration trajectories of a robot for a new task are fed into an instant policy by following a prior LLM-based method (\ie KAT \cite{dipalo2024kat}) to generate the robot trajectory for completing a given task. This process is performed iteratively to gather multiple response trajectories. The robust trajectory is captured from the set of response trajectories by using the Student's t-regression model \cite{lange1989student}, which is a well-established model for ignoring outliers (\ie hallucinations). RIP minimizes hallucinations, thereby enabling a reliable robot agent with a robust trajectory. Validation in both simulated and real-world environments demonstrates that RIP significantly outperforms state-of-the-art imitation learning (IL) methods, particularly in low-data scenarios for everyday tasks.

\section{RELATED WORKS}
\label{sec:rel}
\subsection{Imitation Learning from Human Demonstration}
IL is a promising approach in which robots learn to perform tasks autonomously by observing human demonstrations rather than relying on manual engineering. Recent advances in deep learning architectures have enabled the imitation of a broader range of complex human behaviors. For example, highly expressive architectures such as energy-based models \cite{florence2022implicit} and diffusion models \cite{chi2023diffusion} can efficiently learn probabilistic human behaviors, including discontinuities and multioptimality. Furthermore, training a transformer-based architecture on large and diverse datasets containing various tasks has been shown to lead to generalizable policies for multiple tasks \cite{o2024open, black2024pi_0, kim2024openvla}. However, even when training expressive policy models on extensive datasets, these policies often require additional demonstration data and sophisticated fine-tuning to adapt to new tasks and environments.

\subsection{In-context Imitation Learning for Instant Policy}
ICIL is a notable paradigm that enables robotic policies to adapt instantly to new tasks using a few human demonstrations without explicit policy updates. To this end, deep learning methods, such as contrastive learning, can train a robot to identify similarities between tasks, enabling it to control a robot for test tasks based on similar training tasks \cite{jang2022bc}. Furthermore, the transformer architecture enables similarity matching based on a self-attention mechanism, which led to the proposal of a more concise and highly capable ICIL approach \cite{fu2024contextil}. However, these approaches are effective for tasks similar to those they are trained on; thus, they require sufficient data to cover a vast range of tasks, which is still lacking in the robotics domain. The KAT method addresses this problem by converting robotic data into text, enabling off-the-shelf LLMs \cite{achiam2023gpt} to be reused as policies, yielding performance comparable to advanced IL methods with fewer than 10 demonstrations \cite{dipalo2024kat}. Despite their effectiveness, LLMs face the issue of hallucinations \cite{ji2023hallucination}, where they occasionally provide responses that do not fit the given context, and LLM-based KAT is also vulnerable to this.

\subsection{Mitigating Hallucination in Large Language Models}
Numerous studies have been conducted to mitigate these hallucinations in LLM. Two main streams of research include the following: estimating the uncertainty of LLM predictions by designing explicit probabilistic models \cite{ling2024uncertainty} and estimating the implicit confidence from LLM responses \cite{manakul2023blackbox_hallucination}. The estimation of the implicit confidence from LLM responses is based on the fact that sampled responses are likely to be consistent if the LLM knows a given question, whereas hallucinated responses diverge and contradict each other. This is implemented in a sampling-based approach that assigns a high confidence score to consistent text tokens among the responses sampled from the LLM, and has become more widespread because of its flexibility for diverse applications \cite{campos2024conformal_llm, ren2023robots} and statistical guarantees \cite{vovk2005conformal_prediction}.
However, in robotics, most research has focused primarily on discrete language data \cite{ren2023robots}, which poses challenges in applying these techniques to continuous robot trajectory data. To address this problem, this study proposed an approach that employs a Student's t-regression model \cite{lange1989student} to extract reliable trajectories from the continuous robot trajectory data generated by LLM and investigated its effectiveness on the ICIL of robotics.

\section{FORMULATION}
\label{sec:method}
In this paper, we consider an ICIL scenario in which a human expert provides a few demonstrations of task execution, and a robot immediately imitates these demonstrations to autonomously perform the new task. The expert demonstrations are represented as $\mathcal{D}=\{d_i\}_{i=0}^{I}$, where $I$ is the total number of demonstration episodes, and each demonstration $d_i$ consists of a sequence of observations $o^i$ received by the robot and a sequence of desired actions $a^i$ that the robot should exhibit to complete the task. 
The objective of ICIL is to develop an instant policy $\Phi(\mathcal{D}, o') \rightarrow \hat{a}$ that effectively generalizes the expert policies implicitly expressed in $\mathcal{D}$, enabling appropriate actions $\hat{a}$ to be predicted based on new observations $o'$.

\begin{figure*}[th!]
    \centering
    \includegraphics[width=0.9\textwidth]{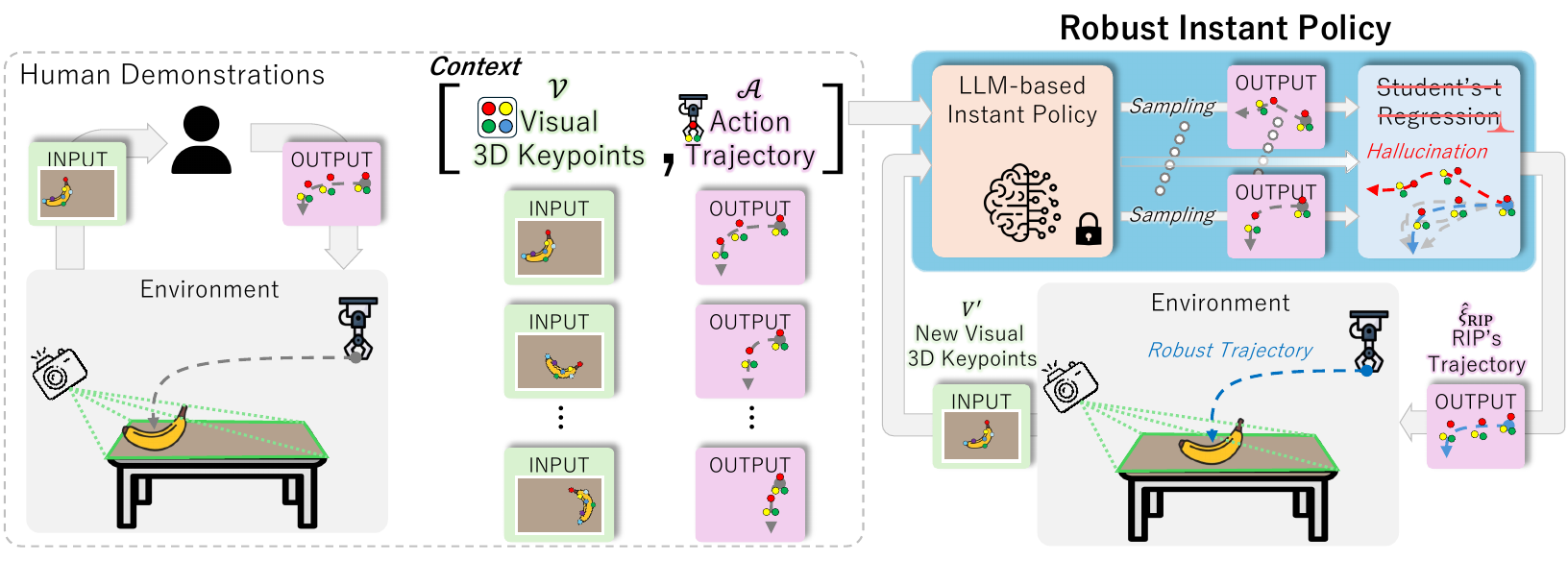}
    \caption{Overview of RIP in a banana-picking task. The input of the initial image and output of the robotic gripper's trajectory are collected through human demonstrations. From the demonstration dataset, contextual text data is tokenized: the visual 3D keypoints with semantic and geometric similarities are extracted from image inputs, and the action trajectory consists of a 3D triplet representing the gripper's posture. The LLM-based instant policy is fed the context data and new image keypoints, sampling the action trajectories multiple times. Using the Student's t-regression model, a robust trajectory is captured from the set of sampled action trajectories, and the robot performs the task following that reliable trajectory.
        }
    \label{fig:overview}
\end{figure*}

\subsection{Leveraging Large Language Models as an Instant Policy} \label{sec:method:KAT}
A key feature of the ICIL is that it does not require additional updates to the policy model parameters. Although certain approaches explicitly train a specific model based on robot data to achieve this feature, KAT show that off-the-shelf LLMs can function as instant policies without fine-tuning \cite{dipalo2024kat}. 
Following this notion, this study was formulated based on the LLM-based approach as follows:

At the beginning of each episode, the robot captures an RGB-D image observation $o^i$. During demonstration $d_i$, the robot collects observations $o^i$ along with a set of action trajectories $\{a^i_t\}_{t=1}^{T}$ of length $T$. When executing an instant policy, the robot records a new observation $o'$ and infers a series of actions $\hat{\xi} = \{\hat{a}_t\}_{t=1}^{T}$ that replicate the expert behavior observed in the demonstration dataset $\mathcal{D}$.

To accelerate the inference capability of LLMs, 3D keypoints-based observation and action space are introduced: 3D visual keypoints extracted from complex RGB-D images as observation and triplet 3D points that can describe the robot's end-effector posture as action.

To extract 3D visual keypoints based on semantic and geometric similarities, a state-of-the-art vision transformer model called DINO \cite{caron2021dino} is used, and the process is as follows:
\begin{enumerate}
    \item Extracting DINO descriptors from RGB-D image $o^i$ in each demonstration yields $z^i \in \mathbb{R}^{N \times 6528}$, with $N$ as the number of patches in each image; that is, each descriptor contains informative features of the patch.
    \item Extracting $K$ descriptors for each image that are most similar to other images by comparing descriptors between images via a nearest neighbor search algorithm \cite{amir2021dino_vit}, from which $K$ 2D keypoints are calculated.
    \item Projecting each 2D keypoints into 3D space by using the depth of each image and known camera intrinsic parameters yields $\mathcal{V} \in \{V_i\}_{i=0}^{I}$, where $V_i=\{v_k\}_{k=1}^{K}$ is a set of visual 3D keypoints $v_k$ of each image.
\end{enumerate}
For a new observation $o'$, we also extract the DINO descriptors $z'$. We then find the $K$ nearest neighbors for each descriptor in $\mathcal{V}$ and extract the corresponding visual 3D keypoint $V'$.

In addition, the action sequence is defined by the trajectory of the robot's end-effector poses. Each unique pose of the robot end-effector is defined by a triplet of 3D points ($\tau^i_t = [p^i_{t,0}, p^i_{t,1}, p^i_{t,2}]$) that represent the $x, y, z$ positions of the gripper and both fingertips when the gripper is open. The state of the gripper is represented by the variable $g^i_t$, where $g^i_t = 0$ indicates that the gripper is open and $g^i_t = 1$ indicates that it is closed. Therefore, action is defined as $\mathbf{a}^i_t = [\tau^i_t, g^i_t] \in \mathbb{R}^{10}$, and all action sequences in demonstration is defined as  $\mathcal{A}=\{\xi^i\}_{i=1}^{I}$, where $\xi^i = \{\mathbf{a}^i_t\}_{t=1}^{T}$. We note that all the 3D coordinates are within the world frame.

At the deployment phase, the text token consisting of a context $[\mathcal{V}, \mathcal{A}]$ and a new visual 3D keypoint $V'$ is input into the LLM, and LLM generates a new trajectory of the desired end-effector motion to complete a task as an instant policy as follows: $\Phi_{\text{KAT}}([\mathcal{V}, \mathcal{A}], V') \rightarrow \hat{\xi}$, where $\hat{\xi} = \{\hat{\mathbf{a}}_t\}_{t=1}^{T}$.
For further details, please refer to \cite{dipalo2024kat}.

Although this instant policy performs favorably compared with other IL approaches, its applicability is not yet stable because of the LLM's hallucinations, which generate action sequences that are considerably different from those inherent in demonstrations $\mathcal{D}$. To overcome this drawback, in the next section, we present a novel approach that captures the uncertainty from an instant policy $\Phi_{\text{KAT}}$ and ignores hallucinations for robust applications.

\section{ROBUST INSTANT POLICY}\label{sec:rip}
In this section, we introduce a novel stochastic approach for enhancing the robustness of LLM-based instant policy. 
We start from the simple notion that if the LLM-based policy knows a given task, the sampled trajectories are consistent, whereas the hallucinated trajectories diverge from each other. Thus, hallucinated trajectories can be recognized as outliers that deviate from a consistent trajectory. 
Therefore, we propose RIP that utilizes the Student's t-regression model \cite{lange1989student} that excels at ignoring these outliers and generates a reliable robot trajectory to complete tasks (\fref{fig:overview}).

Initially, a set of action trajectories $\mathcal{A}'=\{\hat{\xi}^{q}\}_{q=1}^{Q}$ is generated by querying the instant policy $Q$ times using the same text token (\ie context $[\mathcal{V}, \mathcal{A}]$ and a new observation $V'$). 
Notably, in practice, this process is concurrent to avoid computational complexity as $Q$ increases. 
The length of each trajectory can also vary; therefore, the time step $t$ is normalized to the maximum length of each trajectory $T$ to align the generated trajectories based on the start and end of the episode, similar to a previous study \cite{eerland2016timewarping}.
For simplicity without loss of generality, all trajectories have the same length $T$, and the one-dimensional action $\hat{a}_t \in \hat{\mathbf{a}}_t$ is used in the subsequent formalization.

Given a set of trajectories $\mathcal{A}'$ involving hallucinations, the aim is to train an action estimator to approximate a consistent trajectory. 
To this end, we employ the Student's t-regression model \cite{lange1989student} as our action estimator, which is known to enable considerable robustness to outliers than the standard Gaussian distribution \cite{bishop2006prml}. Specifically, our action estimator is defined as $\mathcal{S}_{\theta}(\hat{a}_t|t;\nu)$, which outputs the Student's t-distribution with a mean network $\mu_{\theta}$ and variance network $\sigma^2_{\theta}$ for a given time step $t$ with parameter $\theta$:

\begin{align}
    &\mathcal{S}_{\theta}(\hat{a}_t|t;\nu)\nonumber\\
    &= \frac{\Gamma((\nu+1)/2)}{\Gamma(\nu/2)\sqrt{\nu\pi\sigma_{\theta}^2(t)}}\bigg\{1 + \frac{(\hat{a}_t - \mu_{\theta}(t))^2}{\nu\sigma_{\theta}^2(t)}\bigg\}^{-(\nu+1)/2},
    \label{eq:model}
\end{align}
where $\Gamma(\cdot)$ is the gamma function and $\nu$ is the degree of freedom with a positive real value. $\nu$ is a hyperparameter that regulates the sensitivity to outliers. As it approaches infinity (\ie $\nu \rightarrow \infty$), the distribution resembles a normal Gaussian distribution. Please see details on \cite{bishop2006prml}.

Subsequently, to capture Student's t-distribution of action trajectories, the loss of the action estimator is defined as a negative log-likelihood:
\begin{align}
    \mathcal{L}(\mathcal{S}_{\theta}|\mathcal{A}')= \sum^{Q}_{q=1}\sum^{T}_{t=1}-\log \mathcal{S}_{\theta}(\hat{a}^{q}_t|t;\nu).
\label{eq:loss}
\end{align}
Therefore, our objective function is to train the parameter $\theta$ of the action estimator by minimizing the expected loss along the action trajectories $\mathcal{A}'$ derived from an instant policy.
\begin{align}
    \hat{\theta}=\argmin_{\theta}
    \mathbb{E}_{\mathcal{A}' \sim \Phi_{\text{KAT}}}[\mathcal{L}(\mathcal{S}_{\theta}|\mathcal{A}')].
\label{eq:obj}
\end{align}

Finally, using the learned action estimator $\mathcal{S}_{\hat{\theta}}$, a consistent trajectory is extracted from a set of action trajectories $\mathcal{A}'$. This calculation is performed over time steps using the mean of the learned action estimator.
\begin{align}
    \hat{\xi}_{\text{RIP}} = \{\mu_{\hat{\theta}}(t)\}_{t=1}^{T}.
\label{eq:traj_rip}
\end{align}
A summary of RIP is presented in \aref{alg}.

\begin{algorithm}
\SetKwComment{Comment}{/* }{ */}
\SetKwInOut{Input}{Input}\SetKwInOut{Output}{Output}

\caption{Robust instant policy (RIP)}\label{alg}
\Input{Instant policy $\Phi_{\text{KAT}}$, number of queries $Q$, context $[\mathcal{V}, \mathcal{A}]$, new observation $V'$}
\Output{Robust action trajectory $\hat{\xi}_{\text{RIP}}$}

\For{$q=1$ \KwTo $Q$}{
    Get actions from an instant policy of KAT: 
    $\hat{\xi}^q \sim \Phi_{\text{KAT}}([\mathcal{V}, \mathcal{A}], V')$ \;
    Aggregate action trajectories: $\mathcal{A}' \leftarrow \mathcal{A}' \cup \hat{\xi}^q$\;
}
Learn the action estimator parameter $\hat{\theta}$ by \eref{eq:obj}\;
Get the consistent trajectory $\hat{\xi}_{\text{RIP}}$ by \eref{eq:traj_rip}
\end{algorithm}

\section{EVALUATION} 
\label{sec:eval}
In this study, a novel robust ICIL approach called RIP was proposed to obtain robust instant robot policies that are immediately applicable to novel tasks. Therefore, our evaluation was conducted using simulated and real-world experiments to answer two main questions: 1) How is RIP comparable to state-of-the-art IL approaches? 2) What is the optimal RIP design for maximizing performance?

\begin{figure*}[t]
    \centering
    \includegraphics[width=\textwidth]{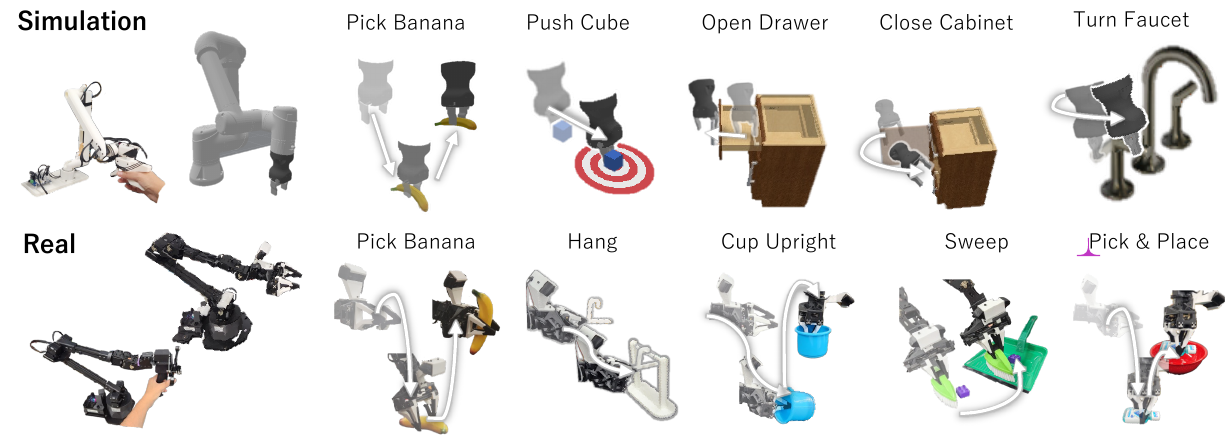}
    \caption{
        Environments and Tasks.
        In both simulated and real-world environments, human demonstrations are collected with leader-follower robot systems, where the movements of a directly human-controlled leader robot are followed by a follower robot with a similar embodiment. Under these robotic environments, RIP and baseline IL methods were evaluated for their capability on 10 everyday manipulation tasks.
    }
    \label{fig:eval:tasks}
\end{figure*}

\subsection{Evaluation Setting}
\subsubsection{Environments and Tasks} \label{sec:eval:tasks}
To evaluate the RIP, as shown in \fref{fig:eval:tasks}, a set of 10 daily tasks was defined for simulated and real-world environments as follows:

\textbf{Simulation environment} was implemented based on the Maniskill3 benchmark \cite{tao2024maniskill3}. The initial RGB-D image of the top view was captured using the built-in Maniskill3 RGB-D camera at the start of each episode. A human provides task demonstrations for a Universal Robotics UR5e 6DOF robot equipped with a Robotiq Hand-E gripper, using a leader-follower teleoperation system called GELLO \cite{wu2024gello}.
These are recorded at \SI{55}{\hertz}, but will be \SI{5.5}{\hertz} during the test phase to ensure stability.
Given demonstrations, the aim is to obtain an instant robotic agent that performs the following tasks:
\begin{itemize}
    \item Pick Banana: A task where a robot picks a banana from a table. A banana is placed randomly; a robot needs to reach it precisely to pick it up.
    \item Push Cube: A task where a robot pushes a cube to the goal position. A cube is placed randomly; a robot agent's nonprehensile capability is evaluated.
    \item Open Drawer: A task where a robot pulls a handle to open a drawer. A drawer is placed randomly; a robot needs to pull a handle correctly to open it.
    \item Close Cabinet: A task where a robot pushes a cabinet door to close it. A cabinet is randomly placed, and a robot needs to push the door in the correct orientation to close it. 
    \item Turn Faucet: A task where a robot turns the faucet. A faucet is placed randomly; a robot needs to turn it in the correct orientation. 
\end{itemize}

\textbf{Real-world environment} was built on the ALOHA \cite{zhao2023aloha} robotic system that also employed leader-follower teleoperation with two 6DOF robot arms. Demonstrations are recorded at \SI{10}{\hertz}, but will be \SI{1}{\hertz} during the test phase to ensure stability. An Intel RealSense D415 camera was mounted on top of the table to capture an RGB-D image at the start of each episode. 
Given the demonstrations, the aim is to obtain an instant robotic agent that performs the following tasks:
\begin{itemize}
    \item Pick Banana: Same as simulation.
    \item Hang: A task where a robot grabs a clothes hanger, reaches, and hangs it on a horizontal stand. A stand is placed randomly; a robot needs to bring a hanger and release it to the correct position.
    \item Put Cup Upright: A task where a robot picks up a horizontally placed cup and places it upright on the table. A cup is placed randomly; a robot needs to grasp and rotate a cup precisely.
    \item Sweep: A task where a robot grabs a brush and sweeps an object into a dustpan. A dust object is placed randomly; a robot needs to sweep it with a brush while touching the table. 
    \item Pick-and-Place: A task where a robot picks up a bottle and places it in a red bowl. A bowl and a bottle are placed randomly; a robot needs to pick a bottle and put it in a red bowl precisely.
\end{itemize}

\begin{figure}[t!]
    \centering
    \includegraphics[width=0.45\textwidth]{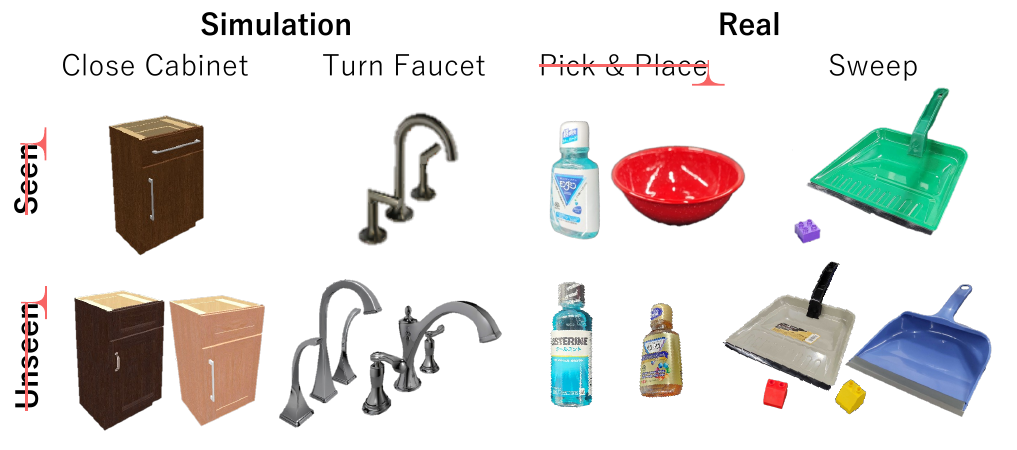}
    \caption{
        Objects used to evaluate the generalizability of unseen objects during training.
    }
    \label{fig:eval:objs}
\end{figure}

Inspired by \cite{dipalo2024kat}, to evaluate the generalizability of objects unseen during training, the following four tasks also test unseen objects in the demonstration: close a cabinet and turn a faucet in a simulation and sweep and pick-and-place in a real-world environment. Specific objects are described in \fref{fig:eval:objs}.

\begin{table*}[t!]
\caption{
    Quantitative Results:
    Comparing the success rate with $10$ demonstrations between RIP and baseline methods for each task.
    The success rate of each method, except KAT in simulation, is measured over $10$ test executions. For KAT in simulation, all five trajectories generated for RIP from one initial image are tested, \ie 50 test executions.
    For each task result, the one in bold is the best.
    Average shows the mean and standard deviation of the results, our method is significantly better than task results marked $^*$ or $^{**}$ (t-test, $p < 5e{-2}$ and $p < 5e{-3}$, respectively).
    }
    \label{table:eval:results}
\centering
\begin{tabular}{p{1.8cm} C{0.9cm} C{0.9cm} C{0.9cm} C{0.9cm} C{0.9cm} C{0.01cm} C{0.9cm} C{0.9cm} C{0.9cm} C{0.9cm} C{0.9cm} C{1.25cm}}
\hline
Methods & \multicolumn{5}{c}{Simulation Tasks Success Rate [\%]} & & \multicolumn{5}{c}{Real-world Tasks Success Rate [\%]} & Avg. [\%] \\ \cline{2-6} \cline{8-12}
    & Pick Banana      & Push Cube          &          Open Drawer   &Close Cabinet & Turn Faucet     & & Pick Banana      & Hang          & Sweep          & Pick \& Place  & Cup Upright   &      \\ \hline
DP                            & $40$             &  $30$             &    $60$          &$\mathbf{100}$      & $70$    & & $60$             & $30$              & $40$               & $10$               & $60$           & $50^{**} \pm 24$  \\ 
KAT-DP                        & $\mathbf{60}$             &  $40$             &    $20$          &$70$       & $60$    & & $\mathbf{100}$             & $50$              & $20$               & $30$               & $\mathbf{70}$           & $52^{*} \pm 24$  \\ 
KAT                           & $48$             & $74$              &         $42$          &$90$     & $78$ & & $80$             & $50$          & $10$           & $30$           & $40$          & $54^{*} \pm 24$  \\ 
$\text{RIP}_\text{Gauss}$ (Ours)    & $40$             & $70$          &      $50$          &$90$      & $\mathbf{90}$ & & $70$             & $60$          & $60$           & $40$           & $\mathbf{70}$          & $64 \pm 17$  \\ 
RIP (Ours)            & $\mathbf{60}$   & $\mathbf{100}$ &    $\mathbf{70}$ &$\mathbf{100}$  & $\mathbf{90}$  & & $\mathbf{100}$   & $\mathbf{90}$ & $\mathbf{70}$  & $\mathbf{50}$  & $\mathbf{70}$ & $80 \pm 17$  \\ \hline
\end{tabular}
\end{table*}

\begin{figure*}
     \centering
     \begin{subfigure}[b]{0.69\textwidth}
         \centering
         \includegraphics[width=\textwidth]{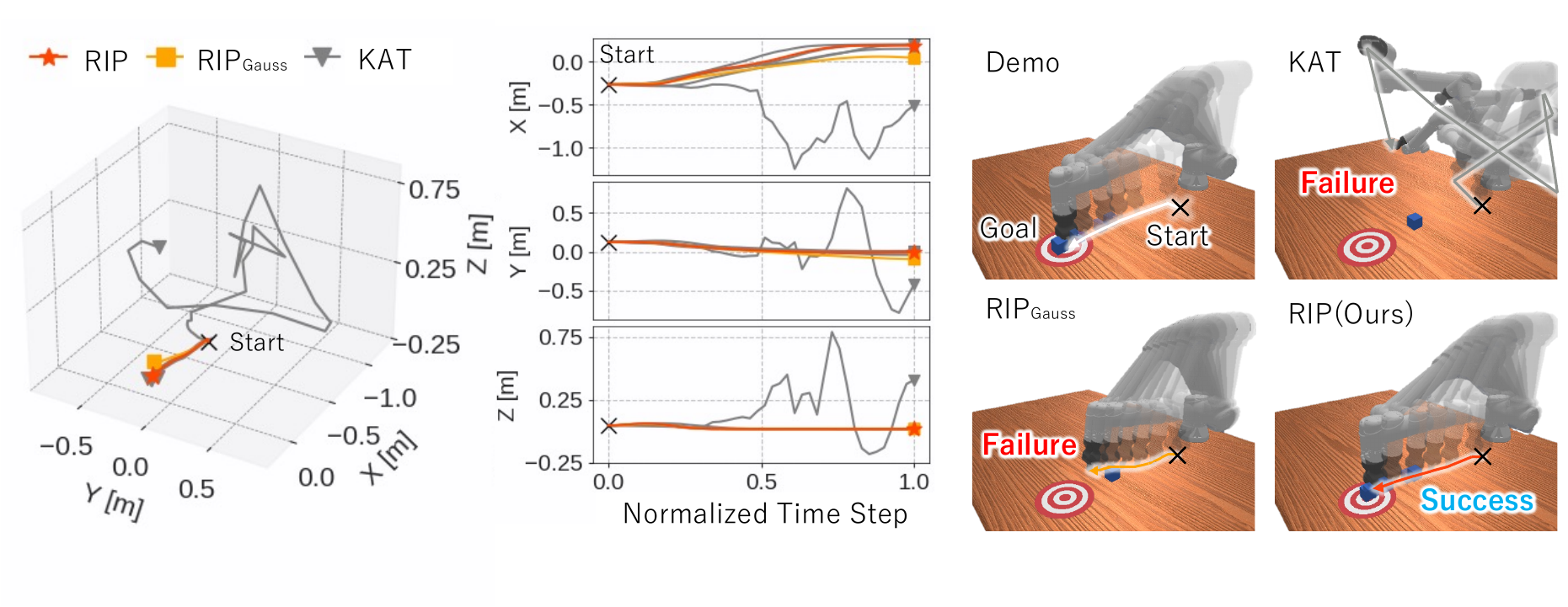}
         \vspace*{-10mm}
         \caption{}
         \label{fig:eval:quali}
     \end{subfigure}
     \hfill
     \begin{subfigure}[b]{0.29\textwidth}
         \centering
         \includegraphics[width=\textwidth]{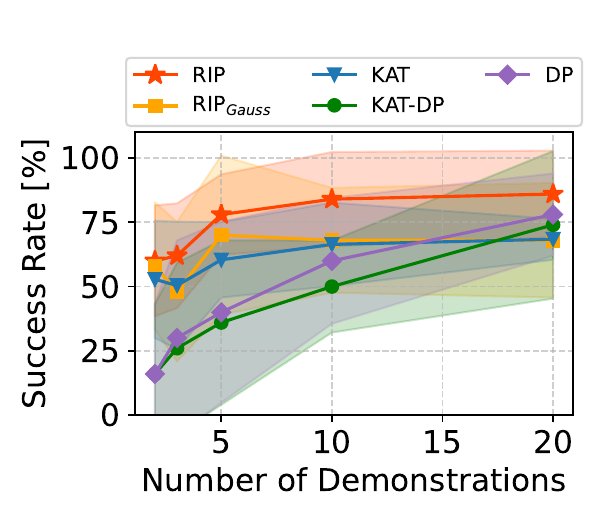}
        \vspace*{-6mm}
        \caption{}
        \label{fig:eval:sim:quanti}
     \end{subfigure}
    \vspace*{-2mm}
    \caption{Evaluation Results: (a) Qualitative analysis comparing trajectories generated by LLM-based instant policies (\eg RIP, $\text{RIP}_\text{Gauss}$ and KAT) on push cube simulation tasks. Note that only the trajectory of the gripper position is represented for intuitive analysis. (b) Quantitative analysis comparing the success rate of RIP and baselines regarding the number of demonstrations. All results represent the mean (line) and standard deviation (shaded) of the success rate on all the simulation tasks. The success rate of each method, except KAT, was measured over $10$ test executions. For KAT, all five trajectories generated for RIP from one initial image were tested, \ie 50 test executions.}
\end{figure*}

\subsubsection{Comparison Methods}
In this evaluation, we compared our method (RIP) with other IL methods, including the following:
\begin{itemize}
    \item Diffusion Policy (DP) \cite{chi2023diffusion}: A state-of-the-art IL algorithm that demonstrated superior learning efficiency and generality compared to several previous IL methods.
    \item Keypoint Action Tokens (KAT) \cite{dipalo2024kat}: A state-of-the-art in-context IL algorithm that introduces an LLM as an instant policy by introducing the keypoint-based observation-action space described in \sref{sec:method:KAT}.
    \item KAT-DP \cite{dipalo2024kat}: An IL algorithm that combines KAT and DP on \cite{dipalo2024kat}. Utilizing the keypoint representation of KAT for the observation-action space instead of the raw image input from the original DP resulted in performance comparable to KAT \cite{dipalo2024kat}.
    \item RIP with Gaussian ($\text{RIP}_{\text{Gauss}}$): A baseline for ablation studies in RIP; it uses a normal Gaussian distribution (\ie $\mathcal{N}(\hat{a}_t|\mu_{\theta}(t), \sigma_{\theta}^2(t))$) instead of the Student's t-distribution.
\end{itemize}
Notably, the keypoint-based method extracts $K = 10$ keypoints from observations recorded at the beginning of the episode as described in \sref{sec:method:KAT}. 
All the LLM-based methods use GPT 4o \cite{achiam2023gpt} as an instant policy model.
RIP uses a fixed number of degrees, $\nu=1.5$, and a query count of $Q=5$. These hyperparameters are chosen to improve the success rate, which is analyzed in \sref{sec:exp:result:q2}. 
See \sref{apdx:dino} for detailed setting of 3D keypoints extraction and RIP model ($S_{\theta}$).

\subsubsection{Downsampling Action Demonstration Dataset}\label{sec:exp:set:downsample}
In practice, the redundant action trajectory length $T$ degrades the performance of LLM-based instant policies \cite{dipalo2024kat}. KAT employs an approach that uniformly down-samples data from a dataset collected at a high frequency to obtain a length of 20--30. However, this downsampling may omit the key steps of the demonstration. In particular, in tasks involving grasping or releasing, downsampling reduces the gripper-action precision of the demonstration. To alleviate this, we first mask the time steps of the start and end of episodes as well as the gripper actions $|g^i_{t+1}-g^i_t|=1$ when it is activated to open or close, and we uniformly sample the actions between these masked data to achieve a length of approximately 30. 
This downsampling was applied to all comparison methods. This technique improved the performance of our method (RIP) in tasks involving gripper actions. The analysis is described in \sref{sec:exp:result:q2}.

\subsection{Results}

\subsubsection{How does RIP comparable to state-of-the-art imitation learning approaches?}\label{sec:exp:result:q1}
Qualitative and quantitative analyses were conducted to answer this question.

\textbf{Qualitative Results:}
The results of the qualitative analysis comparing the trajectories generated by LLM-based instant policies (RIP, $\text{RIP}_\text{Gauss}$, and KAT) are shown in \fref{fig:eval:quali}. 
Most of the trajectories generated by KAT accomplished this task by capturing the context provided by human demonstrations. 
However, one outlier was significantly different from the others, and the robot failed the task, which could be attributed to hallucinations of the LLM.
In the case of $\text{RIP}_\text{Gauss}$, averaging over the trajectories of the KAT, the effect of this outlier cannot be ignored. It generates a trajectory close to the hallucinated trajectory, that is, the X trajectory of $\text{RIP}_\text{Gauss}$ is $14~\mathrm{cm}$ less than that of RIP at the final step. This deviation caused the robot to fail to reach the cube, resulting in task failure.
By contrast, RIP can average the trajectories while ignoring the outliers owing to hallucinations and succeeds in the task.

\begin{figure*}
     \centering
     \begin{subfigure}[b]{0.27\textwidth}
         \centering
         \includegraphics[width=\textwidth]{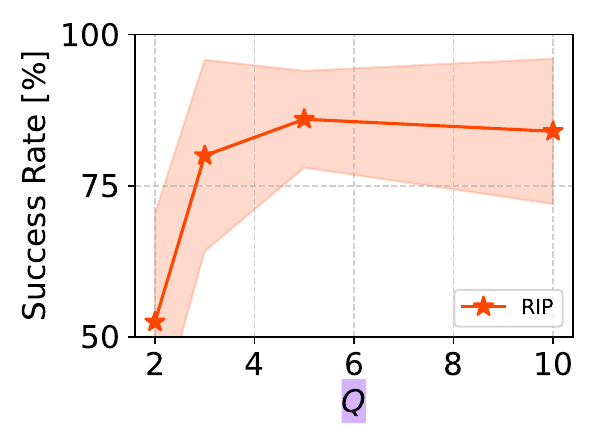}
         \vspace*{-9mm}
         \caption{}
         \label{fig:eval:hyper:q}
     \end{subfigure}
     \hfill
     \begin{subfigure}[b]{0.27\textwidth}
         \centering
         \includegraphics[width=\textwidth]{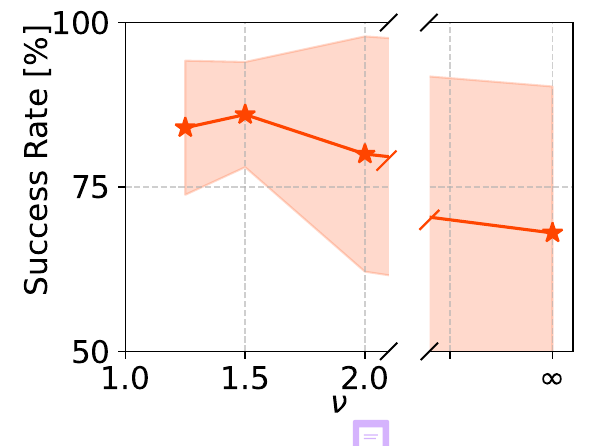}
         \vspace*{-9mm}
         \caption{}
         \label{fig:eval:hyper:v}
     \end{subfigure}
     \hfill
     \begin{subfigure}[b]{0.27\textwidth}
         \centering
         \includegraphics[width=\textwidth]{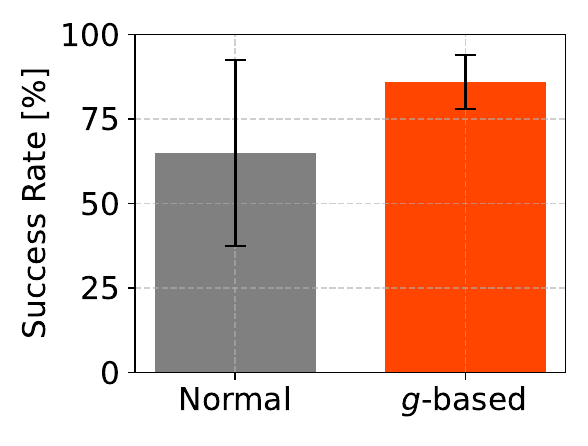}
         \vspace*{-9mm}
         \caption{}
         \label{fig:eval:downsamp}
     \end{subfigure}
    \vspace*{-2mm}
    \caption{Design Analysis of RIP: RIP has three main design elements ($Q$, $\nu$, and downsampling). (a) and (b): Success rate comparison of RIP for different $Q=\{2, 3,5,10\}$, $v=\{1.25, 1.5, \infty\}$, where $\infty$ is the same as $\text{RIP}_\text{Gasuss}$. The results present the mean and standard deviation of the success rates obtained by testing the RIP with $20$ demonstrations $10$ times for each simulated task. (c): Success rate comparison of RIP without and with grasping-action-based downsampling, represented as Normal and $g$-based, respectively.
    The results present the mean and standard deviation of the success rates obtained by testing the RIP with $10$ demonstrations $10$ times for each task.
    }
    \label{fig:eval:hypers}
\end{figure*}

\textbf{Quantitative Results:}
The results of the quantitative analysis comparing the policy success rates of each method are presented in \tref{table:eval:results} and  \fref{fig:eval:sim:quanti}. 

\tref{table:eval:results} lists the policy success rate for each comparison method in a learning scenario where only a limited number of demonstrations ($I=10$) is provided for each manipulation task. Although the LLM-based instant policy methods (KAT, $\text{RIP}_\text{Gauss}$, and RIP) do not require additional training after receiving demonstrations, their average performance outperforms that of the other baseline methods (DP and KAT-DP), which require extra training for each task. Notably, the RIP method achieved a significant improvement in the performance for KAT, with a $26\%$ increase, and showed the best results across all tasks. 
Furthermore, there is a $16\%$ improvement compared with our ablation method ($\text{RIP}_\text{Gauss}$), which aligns with the qualitative results and validates the effectiveness of our Student's t-based design.

In addition, \fref{fig:eval:sim:quanti} shows the quantitative results of simulation tasks that investigate the policy success rate for each comparison method over a wider range of demonstration amounts. The results indicate that up to $10$ demonstrations, the LLM-based instant policy method outperformed the other baseline methods. This finding supports the notion that instant policies can yield higher success rates than state-of-the-art IL approaches, without the need for additional training. However, LLM-based instant policies that lack robustness against hallucinations were surpassed by other baseline methods when the number of demonstrations reached $20$ and failed to achieve a success rate above $70\%$. In contrast, our approach (RIP) consistently demonstrated a superior probability of success across all demonstration amounts. 
Overall, both the qualitative and quantitative results show that our approach (RIP) is more effective than other state-of-the-art IL methods, particularly in scenarios where the number of demonstrations is limited.

\subsubsection{What is the optimal design of RIP to maximize performance?} \label{sec:exp:result:q2}
To investigate this question, we conducted a quantitative analysis of three key factors in RIP design: the number of queries ($Q$), degrees of freedom ($\nu$), and downsampling methods. The results are presented in \fref{fig:eval:hypers}. 

\fref{fig:eval:hyper:q} displays a comparison of the success rate of RIP with varying numbers of queries ($Q$). RIP must extract a robust trajectory from a set containing a sufficient number of reliable trajectories. For $Q = \{2\}$, reliable trajectories may not be obtained sufficiently, making RIP vulnerable to hallucinations, and its success rate is similar to that of KAT. 
By contrast, for $Q \geq 3$, the RIP success rate increases with an increase in $Q$, peaking at $Q=5$. After this point, performance stabilizes, showing no significant improvements; thus, $Q=5$ was used in \sref{sec:exp:result:q1}.

\fref{fig:eval:hyper:v} shows the quantitative analysis of the success rate of RIP for different degrees of freedom ($\nu$). In RIP, the degree of freedom $\nu$ can be interpreted as the level of tolerance of the hallucinated trajectories. When $\nu=\infty$, equivalent to $\text{RIP}_\text{Gauss}$ as described in \sref{sec:rip}, the success rate is the lowest because it is not sufficiently robust against hallucinated trajectories. As $\nu$ decreased, the success rate increased until $\nu=1.5$. We used $\nu=1.5$ in \sref{sec:exp:result:q1} because this setting achieved the highest success rate. 

\fref{fig:eval:downsamp} shows the quantitative analysis of the success rate of RIP according to the downsampling method discussed in \sref{sec:exp:set:downsample}. 
For this evaluation, we focused on tasks in which gripper actions were essential (Simulation: Pick banana; Real: Pick banana, hang, pick-and-place, and cup upright). The use of RIP with gripper-action-based downsampling (success rate: $74\%$) increased the success rate by $22\%$ compared with the method that employs normal downsampling (success rate: $52\%$). This outcome supports our assertion that the data triggering the gripper action may be omitted during the normal downsampling process. Such omissions can lead to less accurate demonstrations and decrease the reliability of instant policies.

\section{DISCUSSION}
Our experiments demonstrate that the proposed method (RIP) significantly enhances the robustness of ICIL, resulting in a reliable robotic instant policy.
Although this study assumed that demonstrators consistently choose a single optimal behavior, in practice, demonstrators often choose multiple optimal behaviors, and certain research attempted to address this challenge in IL \cite{oh2023bdi}. 
Our RIP model currently cannot capture this complexity. Thus, one important direction for future work is to introduce a Student's t-mixture model that can capture multiple optimal trajectories while providing robustness to outliers \cite{peel2000smm}.

Additionally, although our method (RIP) can generate a hallucination-robust trajectory using the Student's t-regression model, it does not explicitly recognize hallucinations. Therefore, the underlying causes of the hallucinations remain unclear. Future work will be directed at identifying the causes of hallucinations in robotic instant policies by introducing an approach for recognizing hallucinations through an explicit LLM probabilistic model \cite{ling2024uncertainty}.

In addition, although the computational complexity remains constant regardless of the number of repeated queries (\ie $Q$) owing to parallelization, current LLMs using transformer architectures still require an output computational complexity of $O(l^2)$ for prompts of length $l$ \cite{vaswani2017attention}. This limitation indicates that current transformer-based LLMs are unsuitable for scenarios that require online control. Therefore, our future work will focus on introducing alternative architectures \cite{gu2023mamba} that exhibit comparable performance but lower complexity.

\section{CONCLUSION} 
\label{sec:conclusion}
This study introduced RIP, a novel ICIL algorithm that leverages stochastic treatment to provide a reliable instant policy for robotic manipulation tasks. 
RIP uses a Student's t-regression model for robustness against the hallucinated trajectories of LLM-based instant policies and captures a reliable trajectory. Our findings demonstrate that robustification achieves state-of-the-art results in IL for various daily tasks.
The design selection analysis reveals the factors that determine the optimal performance of the proposed method. These results show that the RIP allows for a reliable instant robot agent, particularly in scenarios where data are often scarce, such as those in the actual robotics domain.

\section*{APPENDIX}
\subsection{Setting of Extracting Visual Keypoints and Learning RIP}\label{apdx:dino}
As described in \sref{sec:method:KAT}, we utilized the pretrained model provided by Amir et al. \cite{caron2021dino}, known as DINO \cite{caron2021dino}. Although certain parameters (\eg image width (W) and height (H)) were adjusted, as shown in \tref{table:dino}, this did not impact the overall evaluation's generality, and all other parameters remain unchanged from the original.

In addition, the RIP model ($S_{\theta}$) is optimized using \eref{eq:obj} with parameter settings provided in \tref{table:dino}.

\begin{table}
    \centering
    \begin{tabular}{p{1.1cm} C{1.0cm} C{1.0cm} p{1.6cm} C{0.7cm} C{0.7cm}}
         \hline
         \multicolumn{3}{c}{\textbf{Parameters of DINO}} & \multicolumn{3}{c}{\textbf{Parameters of $S_{\theta}$}}\\ \hline
         
         Parameters & Sim & Real& Parameters & Sim & Real \\ \hline
         image W & 100 & 320 & hidden size     & (64,64)   & (64,64)    \\ 
         image H & 100 & 240 & $\nu$              & 1.5       & 1.5        \\ 
         num\_pairs & 10 & 10 & batch size      & 64        & 64         \\ 
         load\_size & 100 & 240 & optimizer       & Adam      & Adam       \\ 
         layer & 9 & 9 & learning steps  & $4\mathrm{e}4$ & $1\mathrm{e}5$ \\ 
         facet & key & key & learning rate   & $1\mathrm{e}{-2}$ & $5\mathrm{e}{-2}$  \\ 
         bin & True & True \\ 
         thresh & 0.15 & 0.15 \\ 
         model\_type & dino\_vits8 & dino\_vits8 \\ 
         stride & 2 & 4 \\ \hline
    \end{tabular}
    \caption{Parameter Setting of DINO. and $S_{\theta}$}
    \label{table:dino}
\end{table}

\section*{ACKNOWLEDGMENT}
This paper is based on the results obtained from a project, Programs for Bridging the Gap between R\&D and the Ideal Society (Society 5.0) and Generating Economic and Social Value (BRIDGE)/Practical Global Research in the AI × Robotics Services, implemented by the Cabinet Office, Government of Japan.


\bibliographystyle{ieeetr}
\bibliography{reference}

\end{document}